\documentclass[letterpaper, 10 pt, conference]{ieeeconf}  % Comment this line out if you need a4paper

\usepackage{graphicx}
\usepackage{color}
\usepackage{amsmath}
\usepackage{url}

\newcommand{\eg}{e.g.\ }
\newcommand{\Reffig}[1]{Fig.~\ref{#1}}
\newcommand{\Refsec}[1]{Sec.~\ref{#1}}

\IEEEoverridecommandlockouts                              % This command is only needed if 
\title{\LARGE \bf
Vision-based Semantic Mapping and Localization for Autonomous Indoor Parking
}

\author{
Yewei Huang$^{3}$, 
Junqiao Zhao$^{*, 1, 2}$,
Xudong He$^{1, 2}$, 
Shaoming Zhang$^{3}$and
Tiantian Feng$^{3}$
\thanks{This work is supported by the National Natural Science Foundation of China (No. U1764261), the Natural Science Foundation of Shanghai (No.kz170020173571) and the Fundamental Research Funds for the Central Universities (No. 22120170232, No. 22120180095).}
\thanks{$^{1}$The Key Laboratory of Embedded System and Service Computing, Ministry of Education, Tongji University, Shanghai
        {\tt\small zhaojunqiao@tongji.edu.cn}}%
\thanks{$^{2}$Department of Computer Science and Technology, School of Electronics and Information Engineering, Tongji University, Shanghai}%
\thanks{$^{3}$School of Surveying and Geo-Informatics, Tongji University, Shanghai}%
\thanks{$^{4}$School of Automotive Studies, Tongji University, Shanghai}%
}

\begin{document}

\maketitle
\thispagestyle{empty}
\pagestyle{empty}

%%%%%%%%%%%%%%%%%%%%%%%%%%%%%%%%%%%%%%%%%%%%%%%%%%%%%%%%%%%%%%%%%%%%%%%%%%%%%%%%
\begin{abstract}

%Autonomous indoor parking without human intervening is one of the most demanded and challenging tasks of the autonomous driving system. 
%The key point to this task is real-time precise indoor localization. 
%However, most indoor parking lots are composed of monotonous texture-less scenes and thus, are hostile to traditional visual feature-based SLAM methods.
In this paper, we proposed a novel and practical solution for the real-time indoor localization of autonomous driving in parking lots.
High-level landmarks, the parking slots, are extracted and enriched with labels to avoid the aliasing of low-level visual features.
We then proposed a robust method for detecting incorrect data associations between parking slots and further extended the optimization framework by dynamically eliminating suboptimal data associations.
Visual fiducial markers are introduced to improve the overall precision.
%Their number and distribution are also analyzed and compared.
As a result, a semantic map of the parking lot can be established fully automatically and robustly.
We experimented the performance of real-time localization based on the map using our autonomous driving platform TiEV, and the average accuracy of 0.3m track tracing can be achieved at a speed of 10kph.

\end{abstract}

%%%%%%%%%%%%%%%%%%%%%%%%%%%%%%%%%%%%%%%%%%%%%%%%%%%%%%%%%%%%%%%%%%%%%%%%%%%%%%%%
\section{INTRODUCTION}

Autonomous driving has been witnessed considerable progress in recent years; the breakthrough has been made in several harsh fields, including obstacle detection, real-time motion planning and high precision localization (mostly based on differential GNSS).
Recently, testing self-driving car can already drive safely in urban and suburban areas\footnote{https://waymo.com, http://archive.darpa.mil/grandchallenge/}.
However, parking in a large indoor parking lot without human intervention is still an unsolved problem.
One critical reason is the lack of robust high precision localization mean in these GNSS forbidden areas.
Traditional indoor localization methods require pre-equipped sensors, such as WiFi, Bluetooth or UWB. 
Wireless signal suffers from occlusion and decays while user's distance to signal sources increases, so a significant number of stations are needed for stability, let alone their relative low precision\cite{stojanovic2014indoor}. 
Laser-based SLAM (simultaneously localization and mapping) system is eligible for localization an unmanned vehicle in environments such as a factory or a warehouse\cite{hess2016real}.
However, this range based representation is of high data volume and is vulnerable to dynamic scenes.
As a result, visual SLAM (VSLAM) built on low-cost cameras became one of the most favorable localization methods.

VSLAM is known to be effective in texture-rich environment\cite{Mur2017ORB}.
Nevertheless, they can easily fail in a monotonously textured scene such as an indoor parking lot.
\cite{Grimmett2015Integrating} adopted sparse feature point based SLAM method with panorama images to localize a car in parking lots.
But the extracted sparse feature can be unstable when the ground floor is stained with tire markings or water spots.
The distortion presented in the stitched panorama images can also disturb the feature extraction.

The direct methods estimate camera poses directly based on photometric error derived from the whole image, thus are more robust than sparse methods in less-textured area \cite{Engel2014LSD,Forster2014SVO}.
HorizonAD applied such a method for indoor parking\footnote{https://github.com/HorizonAD/stereo\_dso}.
However, these methods often require high frame rate and are susceptible to global illumination change, which restricts their usage in unevenly illuminated indoor parking lot \cite{Younes2016ASO}. 
Most importantly, the re-localization based on a pre-built dense map is not trivial since illumination can vary during revisiting.
Therefore, most direct VSLAM methods are rather visual odometries\cite{Engel2014LSD}.
As a result, more stable and legible visual landmarks which are immune to various illuminate condition are demanded.

As a typical kind of semantic landmarks in parking lots, parking slot is now a favorite for researchers
\cite{Houben:2015hq,Grimmett2015Integrating,doi:10.1177/1729881417720781}.
Recently, the deep learning-based method shows its capability of accurate and robust detection of such kind of meaningful objects \cite{Li2017VisionbasedPD}. 
Inspired by these methods, we present a robust VSLAM system based on the recognition of high-level landmarks for parking, i.e., parking slots and their IDs. 
Visual fiducial markers are introduced for improving overall accuracy and robustness. 
Facing the visual aliasing problem of parking slots, we proposed a robust outliers detection and elimination strategy in the optimization stage.
Finally, a two-dimensional map of parking slots can be robustly established which is distinguished from the traditional feature-based or point-cloud map for its stability, re-usability, lightweight and human readable.
Our system is implemented on an autonomous driving vehicle and tested in real parking lots.

Our contributions are: 
\begin{itemize} 
\item We design a practical mapping and localization system using slots and their IDs, which are typical semantic landmarks in the indoor parking lot; 
\item An approach to associate parking slots and their IDs using robust SLAM back-end is proposed;
\item Visual fiducial markers are introduced in parking slot lacking areas as an aid.

\end{itemize}

\section{RELATED WORKS}

%brief review of SLAM
SLAM has long been a classic topic in the robotics field\cite{Cadena:2016fp} and recently became heated in the autonomous driving since many drivable areas are GNSS denied\cite{Bansal-2015-5905}. 
%\cite{Davison2003Real,Davison2007MonoSLAM,Civera20101} use Extended Kalman Filter(EKF) to simultaneously optimize the sensor and landmarks' positions in real-time. 
%However, the computation grows quadratically with the number of landmarks\cite{Bailey2006Simultaneous}.
%Probability filters have many further extensions such as UKF \cite{martinez2005unscented}, Information filter\cite{thrun2005multi}, particle filter\cite{montemerlo2007fastslam, montemerlo2002fastslam} etc.
Filter-based methods\cite{Davison2003Real, thrun2005multi, montemerlo2007fastslam} use probability filters to simultaneously optimize the sensor and landmarks' positions in real-time. 
%However, they all assume the conditional independence of the current measurement with the historical states, which restricts the SLAM into a predict-update iteration loop.
%\COMMENT{JOHN to be more rigorous}
To relax the assumption of conditional independence of the current measurement with the historical states, 
 %Inspired by early researchers on the sparse graph, a 
factor graph-based optimization framework (known as Graph SLAM) was proposed \cite{lu1997globally}.
% This method is closely related to a Markov random field model, thus can involve the influences of all historical measurements, which together with 
Its flexibility together with its accuracy enables the Graph SLAM became the most popular SLAM method\cite{strasdat2010real}.  

% review of VSLAM
Generally, VSLAM methods fall into two groups, so-called feature-based methods (the indirect methods) and direct methods. 
%To emancipate tracking from the map-making procedure probabilistically, PTAM\cite{Klein2007Parallel} separates tracking and mapping into two sub-tasks.
%Keyframes are extracted and used in optimization, but the system cannot deal with large loop closures, it can only handle tasks in small areas(a desk or a room corner). 
%\cite{Lategahn2012City} and \cite{Lategahn2014Vision} build city-scale sparse maps using G2O, a graph optimization framework\cite{K2011G2o}. 
%Nevertheless, the optimization of the sparse 3D point cloud is time-consuming, so the off-line mapping procedure is a must. 
As an example of feature-based methods, ORB-SLAM \cite{Mur2017ORB} offers a stable and efficient graph-based VSLAM system.
With the keyframe detection and the BoW-empowered fast loop closure detection, it performs well in various indoor and outdoor environments. 
However, as low-level features are treated as landmarks in feature-based systems, ORB-SLAM is still easy to fail in texture-less environments.
%
%Because feature-based methods are only capable of creating a sparse feature-based mapping, they cannot be directly used in applications where full reconstruction is demanded, e.g. AR or structure from motion.
To satisfy those applications where full reconstruction is demanded, direct methods based on photometric error and utilize all image pixels are proposed \cite{Engel2014LSD}.
%Comparing to sparse feature-based methods, direct methods can output a semi-dense point cloud with higher quality at a real-time speed. 
But in practice, direct methods require a high rate of overlapping between consequent frames, and the high frame rate is also a necessity since brightness consistency is crucial to estimate the depth accurately.
%Direct methods are also known to be vulnerable to motion blur, camera defocus and global illumination changes \cite{Newcombe2011DTAM}. 
SVO \cite{Forster2014SVO} and DSO \cite{engel2017direct} combine advantages of feature-based method and direct method, and runs extremely fast. 
However, lacking loop closure detection, these odometric methods drift as time increases.
	
% Review of Semantic slam

Traditional SLAM methods do not incorporate humanly understandable meanings (semantics) associated with landmarks into the method, which now is recognized to be crucial for construct a human-readable map and strengthen the descriptive power of landmarks \cite{Cadena:2016fp}. 
\cite{DBLP:journals/corr/LiB16d} added semantic labels to an LSD-SLAM framework to construct a dense map with classes attached to geometric entities, but semantic labels help little in the optimization or localization stages.
SLAM++ \cite{Salas2013SLAM} and Semantic Fusion \cite{Mccormac2017SemanticFusion} employed semantic labels in the RGBD SLAM framework to aid the loop closure. 
However, both methods work in restricted indoor domains, \eg{households or offices}, because of RGB-D cameras' limitations in depth measurement.
%In \cite{Wang2015Lost}, shop names and shop facades are recognized as labels in large indoor shopping spaces.
%\cite{Grimmett2015Integrating} and \cite{Himstedt2017Online} reconstruct the metric map and the semantic map of a parking lot, but they are not used in SLAM but helps the later route planning and parking task.
%\COMMENT{!!semantics in VW paper(2015) is not used in SLAM at all, not sure about the second one }

%\COMMENT{JOHN!!should be very clear about the defects of these methods, i.e. 3D objects? not precise regarding the localization of objects? can be confusing when multiple objects are searched?}

% conclusion of the review part
In a short conclusion, existing VSLAM methods generally could not perform robustly in a texture-less area like an indoor parking lot.
Therefore, more descriptive landmarks, especially landmarks attached with semantics should be used.

%Due to the high false positive rate of object detection, robust data association for semantic landmarks is an urgent need.
%Some methods\cite{sunderhauf2012switchable, latif2013robust, agarwal2013robust, olson2013inference} have offer alternatives for robust loop closures, all of which are designed for pose graph and are not suitable for landmark-based data fusion.

\section{APPROACH}

\begin{figure*}
\centering
\includegraphics[height = 2in]{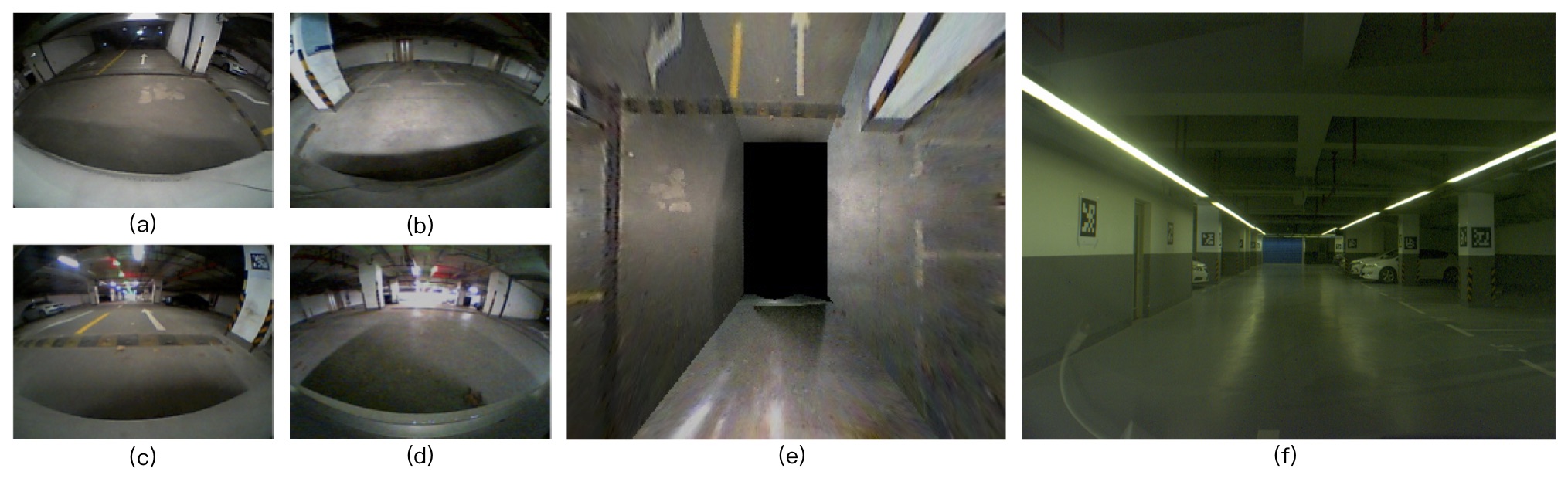}
\caption{Images from fisheye cameras. 
(e) shows the top-view image fused from (a) left (b) right (c) front (d) back . The image from the monocular camera is (f). 
}\label{surround view example}
\end{figure*}

Our semantic VSLAM system includes four fisheye cameras and one monocular camera. 
Four fisheye cameras are fixed at two reflectors, and at the front and rear bumpers, which consist a surround-view system. 
A top-view image is then fused from the surround-view inputs after intrinsic and extrinsic calibration, as shown in \Reffig{surround view example}. 
In the top-view image, which indicates ground textures, parking slots are detected. 
The monocular camera is installed to the left of the rear-view mirror to capture front-view scenes. 
The steering wheel angle, as well as the vehicle speed and heading direction collected by IMU, are also used in our system.

\begin{figure*}
\centering
\includegraphics[height = 1.4in]{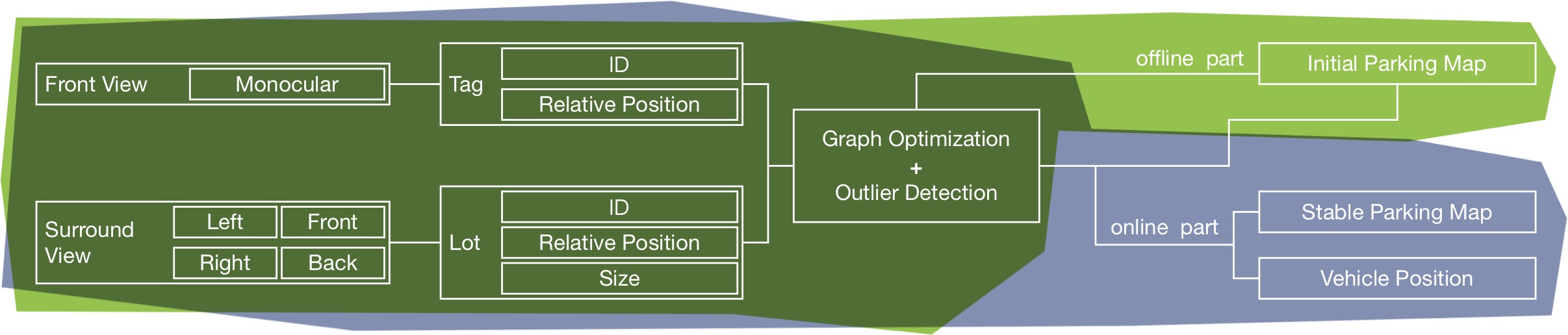}
\caption{Pipeline of the method}
\label{overall_pipeline}
\end{figure*}

%

%We choose parking slot as the landmark since parking slots are the most distinct objects in a parking lot, and the precise locations of parking slots are informative for localization and navigation during autonomous parking.
Our parking slot detector is based on \cite{Li2017VisionbasedPD}, in which corner points of parking slots are detected and assembled(\Reffig{detect slot}).
%Afterwards, parking slots are assembled according to the image patterns around corner points (\Reffig{fig of parking slots assemble}).
Although the CNN-based method is capable of detecting most kinds of corner points fast and robust, the exact shape of the parking slot cannot be known due to the limited visible range of the surround vision system.
As a result, the parking slot can only be guessed initially and we have to optimize the shape of the parking slots in the SLAM system.
Furthermore, the ID of each parking slot should be detected for facilitating data association between parking slots, which will be elaborated in \Refsec{sec:recognition}.

Another kind of landmark used in our system is the visual fiducial marker.
Fiducial markers are introduced as an aid for the constancy of localization since few parking lots are detected near the entrances and exits. 
%In our study, we found fiducial markers may not be fully prohibited but their number can be limited to a small amount. 
We select AprilTags as fiducial tags for its robustness and high-efficiency \cite{Olson2011AprilTag}.

\subsection{CNN based Parking slot Recognition}
\label{sec:recognition}
\begin{figure}[htbp]
\centering
\includegraphics[height = 2.6in]{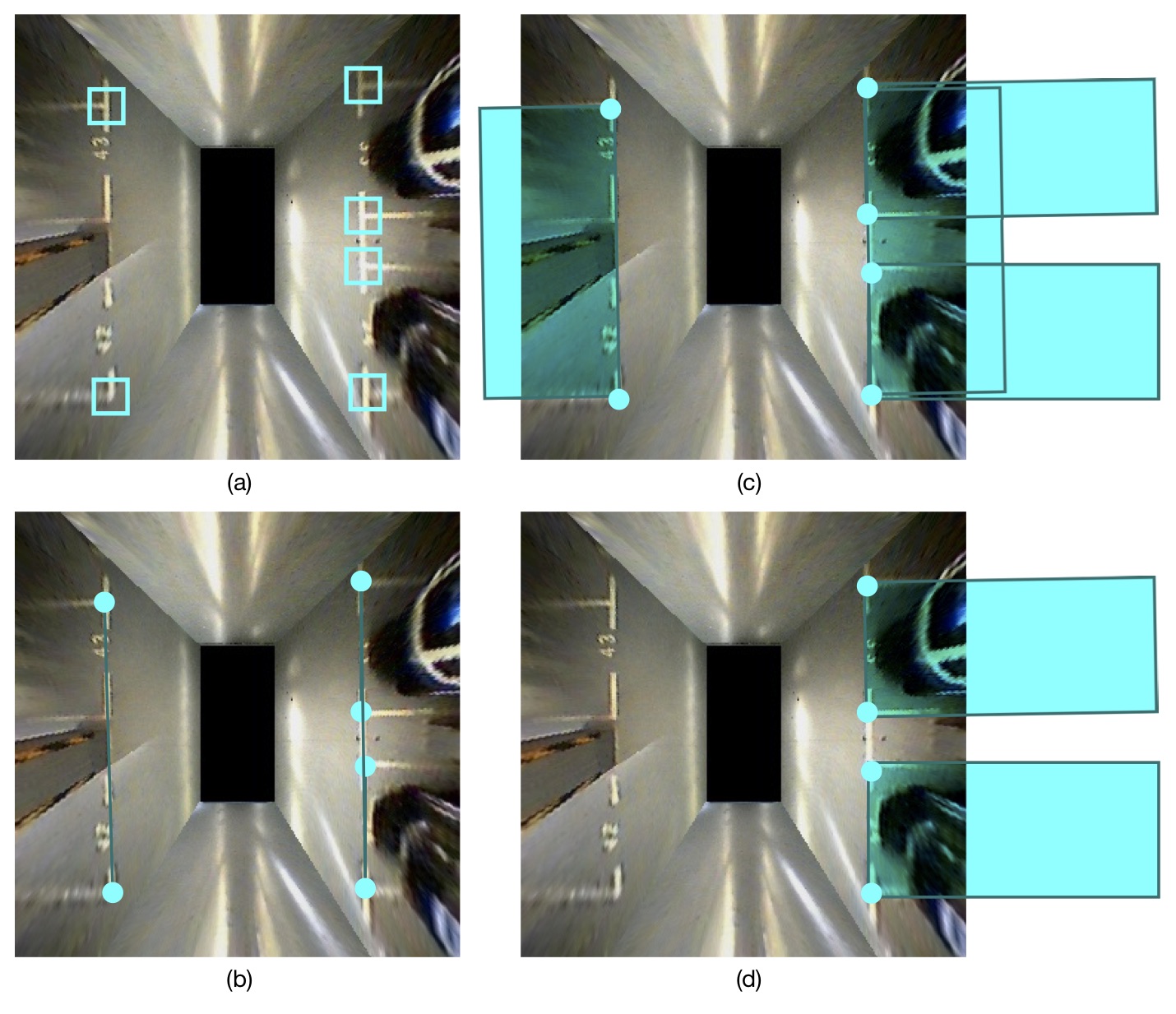}
\caption{
This figure illustrates how parking slots are detected. Corner patterns are detected (a) and assembled (b), which enables the initial parking slot hypothesis (c). Finally, false positives are discarded by constraints based on prior knowledge.
}\label{detect slot}
\end{figure}

We adopt the method proposed by \cite{Li2017VisionbasedPD} to detect parking slots.
It is a CNN-based slot detection method who detect parking slots from calibrated top-view images. 
Slot detection is achieved by firstly recognize the corner patterns from the image.
\Reffig{detect slot}(a) illustrates the examples of detected corner patterns.
Since all the corners of a parking slot may not be entirely observable, the parking slots are estimated according to their entrance-lines (\Reffig{detect slot}(b)), which are determined by the configuration of patterns.
Several constraints are applied to robustify the detection result.
The entrance-line candidate who contains more than two corner patterns is removed to avoid repeated detection.  
Extremely large or small candidates are also discarded since all slots are around the same size.

A slots' shape and direction can serve the further parking and obstacle-avoiding task.
Hence, the precision of a slot's shape and direction are as crucial as that of its position.
To optimize all of them, an additional rectangular constraint is added.%(\Reffig{rectangular constrain})
A parking slot is represented by four landmark points.
Each point is connected with other three by a rectangular constraint,  according to an angular constraint of high confidence and a distance constraint with relatively lower confidence.
The slot is then connected to the global map as an entirety.

\begin{figure*}
\centering
\includegraphics[height = 2.3in]{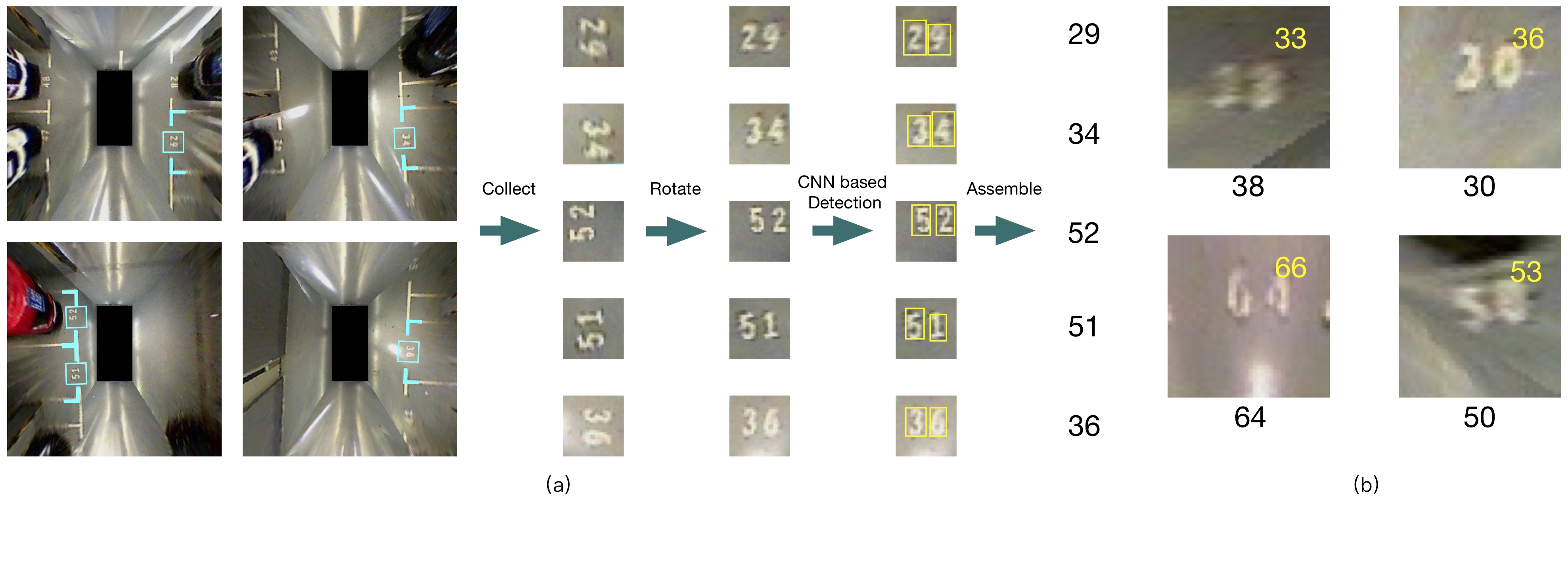}
\caption{
%\COMMENT{add captions of left and right}
(a): the slot ID detection pipeline, (b): examples of harsh image patches for ID detection, digits under images are the true IDs while yellow digits are the detection outputs.
}\label{detect ID}
\end{figure*}

IDs of parking slots are essential for the association of this semantic landmark.
We fine-tuned PVANet to detect each digit in one slot ID \cite{DBLP:journals/corr/HongRKCP16}.
Slot IDs have their fixed position, so entrance-lines of parking slots help locate IDs roughly (\Reffig{detect ID} (a)).
Then image patches containing slot IDs are extracted and detected.
Unfortunately, due to the distorted and blurred texture in the surround view image, even the sophistic detection network could not offer the satisfactory performance.
So we devised a semantic-assisted association method to cope with the uncertainty, which will be detailed in \Refsec{sec:optimization}. 

\subsection{Visual fiducial tag localization}

The surround view system offers an intuitive ground observation at a high frame rate, but the visibility range and resolution of the top-view image are far from satisfactory, limiting the slot detection performance.  
And the calibration for image fusion becomes inaccurate as time goes, which will deteriorate the slot measurement.

Recalling the goal of our practical localization system for autonomous driving and parking in parking lots, certain numbers of faithful landmarks such as visual fiducial markers still have to be incorporated.

We adopted AprilTag and employed the detection framework from the AprilTag C source Open Library \footnote{https://april.eecs.umich.edu/wiki/AprilTags} \cite{Olson2011AprilTag}.
Moreover, the relative position between fiducial tags and the vehicle are solved by the PnP model \cite{Hartley2003Multiple} in a fast and accurate way. 

\begin{figure}
\centering
\includegraphics[height = 1.9in]{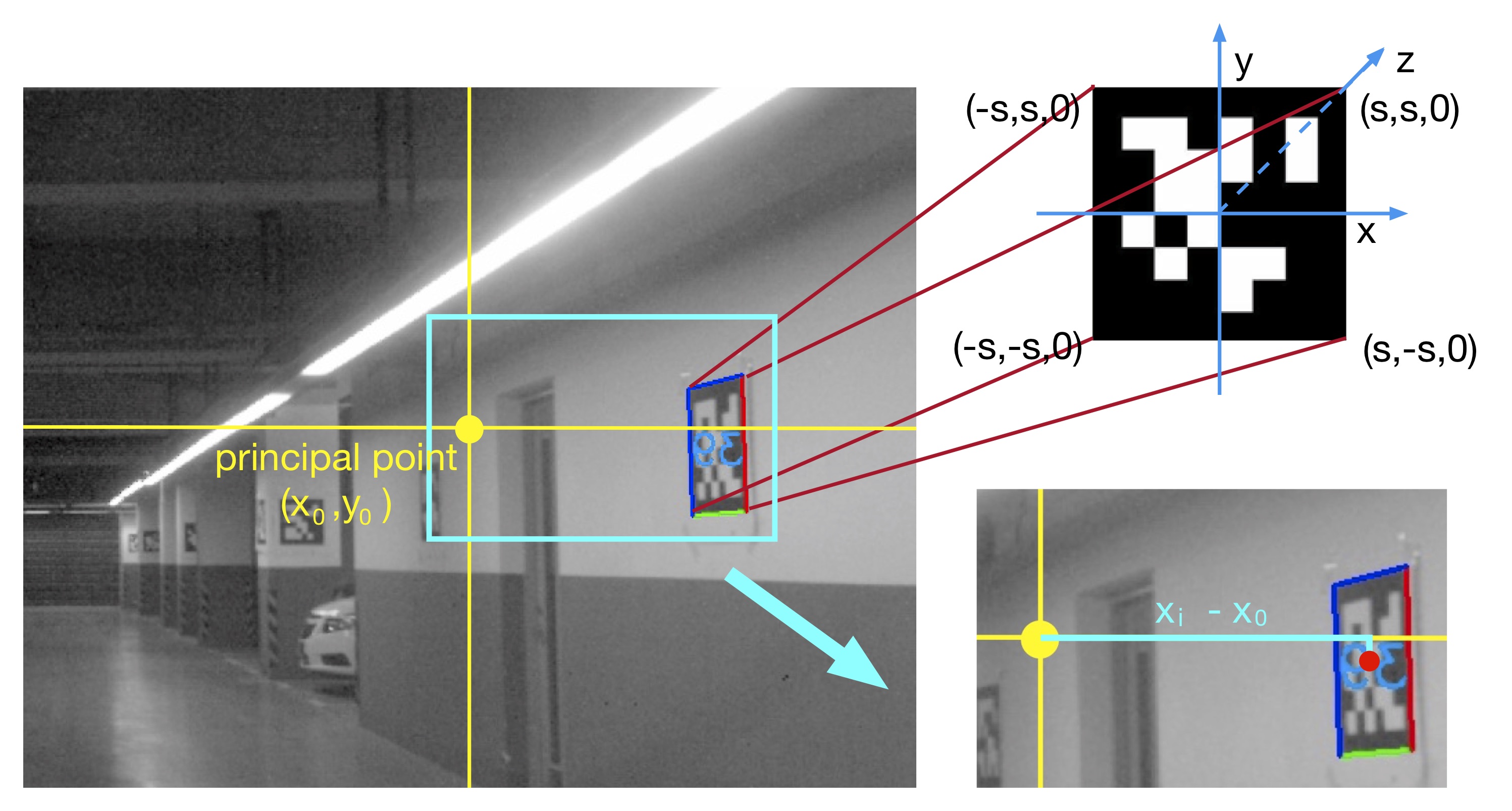}
\caption{
The upper graph shows the match between the hypothetical 3D tag coordinate and the real-world 3D coordinate, while the lower part illustrates how to solve the angle directly.
}\label{solve tag}
\end{figure}

In the PnP model, we assume four corner points of the tag always lie on the same plane, and thus definite the hypothetical 3D tag coordinate(\Reffig{solve tag}).%: whose origin is a tag’s geometrical center, x-axis and y-axis are both parallel to tag’s edge, pointing rightward and upward respectively, and z-axis is perpendicular to the tag plane, pointing inward). 
The relationship between the tag corner in the image and that in the hypothetical 3D tag coordinate can be described as

\begin{center}
${R}_{3 \times 3} \cdot {{K}_{3 \times 3}}^{-1} \cdot {x}_{3 \times 1}+{t}_{3 \times 1}={X}_{3 \times 1}$,
\end{center}

where ${R}_{3 \times 3}$ and ${t}_{3 \times 1}$ are the relative rotation and translation vector, 
${K}_{3 \times 3}$ is the intrinsic camera matrix, 
${X}_{3 \times 1}$ is the coordinate in the hypothetical 3D tag coordinate and ${x}_{3 \times 1}$ is the homogeneous coordinate in the image plane. 

%figure:solve_tag: how tag coordinate and real-world coordinate is accorded, how PnP angle is solved and how the angle is solved by our direct method.

In PnP method, the solution is not always reliable since the initial value greatly affects the iteration result.
So we also get the angle directly from the calibrated image and compares the value with ${R}_{3 \times 3}$ to evaluate whether PnP method works.
As shown in \Reffig{solve tag}, the angle $\alpha$ can directly be calculated by

\begin{center}
  $\alpha = \arctan((x_i - x_0) / f)$, 
\end{center}

where $x_i$ and $x_0$ denote the x coordinate value of the tag center and the principal point respectively, and $f$ is the focal length.
The distance $d$ is the 2-norm of ${t}_{3 \times 1}$ . 
So the tag locates at $x=\sin(a) \cdot d, y=\cos(a) \cdot d$ in the vehicle relative coordinate.

These visual fiducial markers are flexible and easily implemented.
They brought another benefit for the autonomous parking purpose; those fiducial markers can easily indicate the existences of pillars and walls which can only be robustly detected by expensive laser scanners.
This obstacle information can facilitate the route planning inside of a parking lot.

\subsection{Optimization}
\label{sec:optimization}
\subsubsection{Optimization Framework}
We adopt a Graph-based optimization back-end \cite{K2011G2o}.
However, due to fallible detection of parking slots and their IDs from low-quality surround vision images, ambiguities will be presented during data association, which significantly affect the mapping and localization.
Thus, the correct association should be ensured and wrong ones should be detected and discarded in the optimization.
These are performed at both the front-end and the back-end.
At each frame, parking slot observations are pre-associated through their IDs and the nearest neighbor search.
The nearest neighbor is based on the relative offsets between landmarks, which is derived from a Kalman-based extrapolation with the steering wheel, car speed and the compass readings from a cheap IMU as the inputs.

\begin{figure}
\centering
\includegraphics[height = 1.2in]{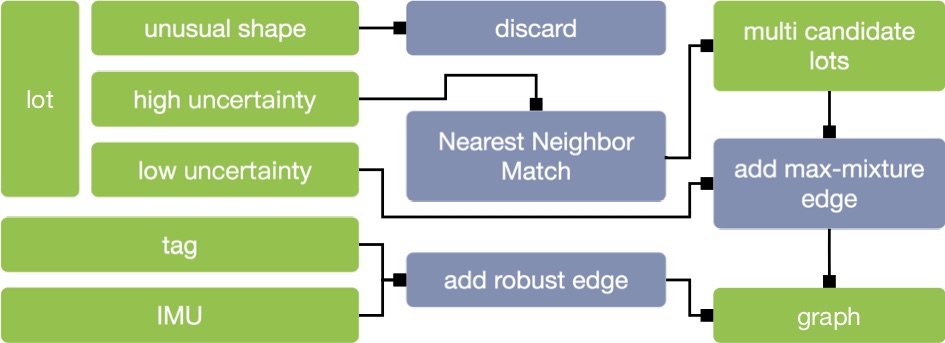}
\caption{
The overall pipeline of optimization
}\label{optimize pipeline}
\end{figure}

We further added the pre-associated landmarks into our graph model based on a Max-Mixture model \cite{Pfingsthorn2014Representing}.
% is therefore introduced and improved to not only detect but also correct mistaken associations.
%And 
The detailed optimization pipeline is shown in \Reffig{optimize pipeline}. 

%\subsubsection{Optimization Framework}
%We adopted G2O \cite{K2011G2o} in our method.
%The least square problem of graph optimization can be expressed by the following equation:
%\[
%F(x)=\sum\limits_{k \in C}e_k{(x_k,z_k)}^T{\Omega}_k e_k(x_k,z_k)
%\]
%\[
%x^{*} = \mathop{argminF(x)}\limits_x
%\]
%where $C$ denotes the total ID group; 
%$x_k$is the parameter block of the $k$Th vertex(containing the vertex’s location and direction), $z_k$ and $\Omega_k$ is the mean and covariance of the $k$Th edge; 
%$e_k(x_k,z_k)$ measures how well $x_k$ matches $z_k$. 
%These elements consist the global error function $F(x)$ , and $x^*$ is the global optimal solution.

%We can get $x^*$ by solving the linear function iteratively, and the uncertainties of $z_k$ are described by noise models who are uni-model Gaussian. 
%This classical graph framework works quite well when there are only fiducial tags and IMU measurements in parking map, but it gives bad results once parking slots are added.
%Since every parking slots may be wrongly associated or detected, it's impractical to measure their noises by a single model. 

%

\subsubsection{Outliers elimination using Max-Mixture Model}
Classical graph optimization method using uni-model Gaussian is sensitive to outliers and fails when there are wrongly associations in graphs.
Several robust methods \cite{sunderhauf2012switchable, latif2013robust, agarwal2013robust, olson2013inference} have been proposed solely for pose graph. Hence they can not be directly used in our landmark-based method.
%Among them, Max-Mixture method is chosen because it's more versatile for landmark association. 
However, after several modifications, the Max-Mixture method can help not only eliminate errors in revisiting landmarks, but also correct wrongly associated slot IDs.

Max-Mixture describes observation ambiguities with multi-model Gaussians. Therefore, wrongly associated data can be suppressed by other mixture elements. 
The Likelihood function of landmark $x$ is expressed by a max-mixture of Gaussians \cite{olson2013inference}:

\begin{center}
$p(z_i|x)=\max\limits_{j}w_j N({\mu}_j , {\Lambda}_j^{-1})$
\end{center}
where $N({\mu}_j , {\Lambda}_j^{-1})$ and $w_j$ denotes the Gaussian distribution and weight of the $j^{\rm th}$ observation $z_j$. 
%To simplify the problem, the sum is substituted by a max operator:

%
In this paper, semantics attached to landmarks, the slot IDs, are used to evaluate slot observations' uncertainties.
A map-scale nearest neighbor search offers candidates for slot observations with highly uncertain IDs. 
Partially detected slot IDs (only one in two digits is recognized) also provide data associating alternatives. 
Afterward, all candidates are added to the factor graph, and only the "max candidate," who has the minimum residual is reserved.
Slot observation with neither high confidence ID nor nearest neighbor candidates is a potential new slot candidate and will be "lazily" added to the map.

\section{EXPERIMENT}

\begin{figure}
\includegraphics[height = 3.4in]{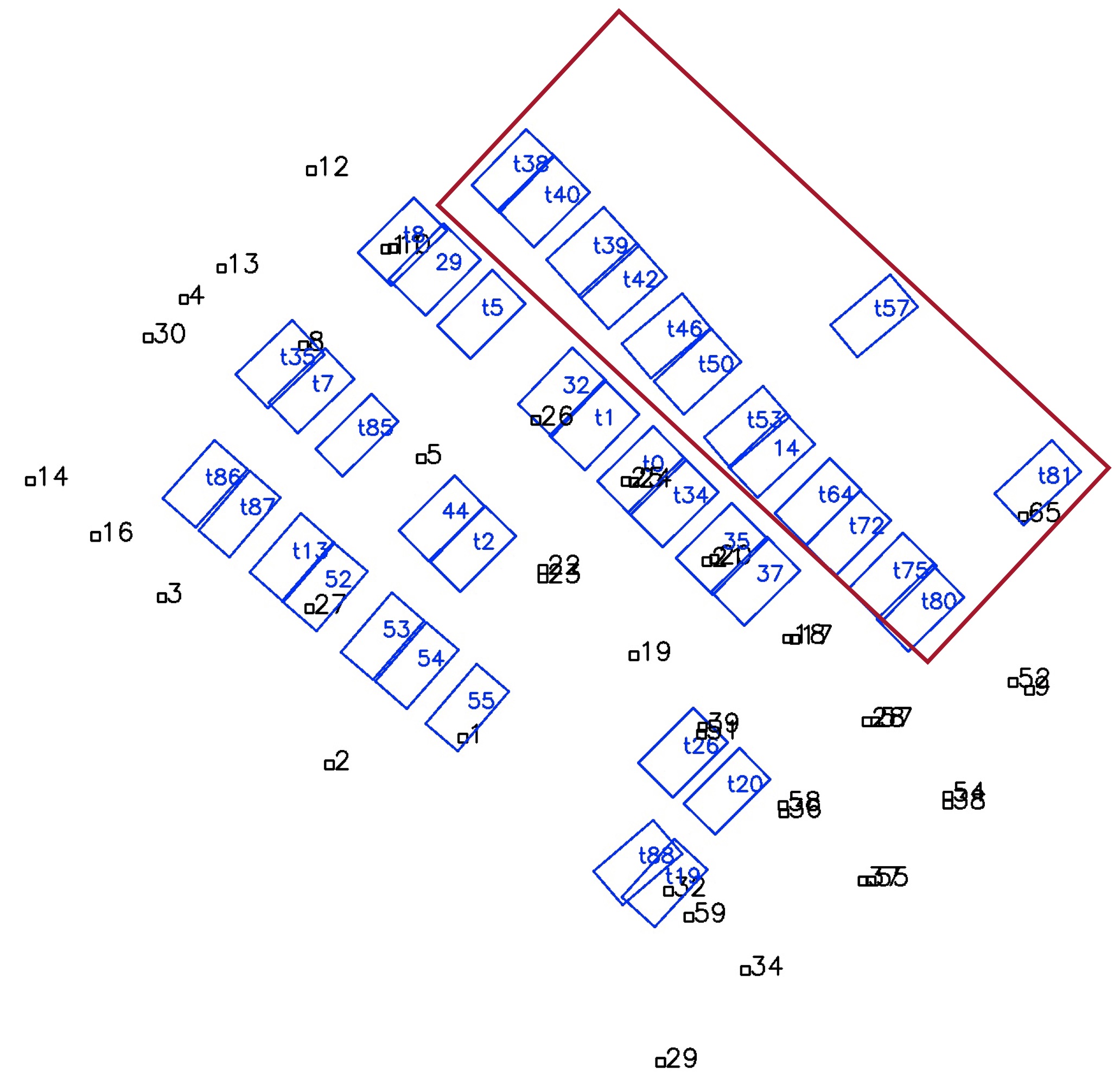}
\centering
\caption{
%\COMMENT{unified the sizes of characters and remove (a)}
%(a)(b) are both maps generated from the same dataset at different time stamp. In both maps, slot IDs starting with "t" are temporary slot IDs for slots where ID detection fails. The replicated slot,t83, in (a) is fused with its nearest neighbour, t75, in (b).
In the map above, lot IDs starting with "t" are temporary slot IDs for slots where ID detection fails.
}\label{map 2}
\end{figure}

\begin{figure}
\includegraphics[height = 3.3in]{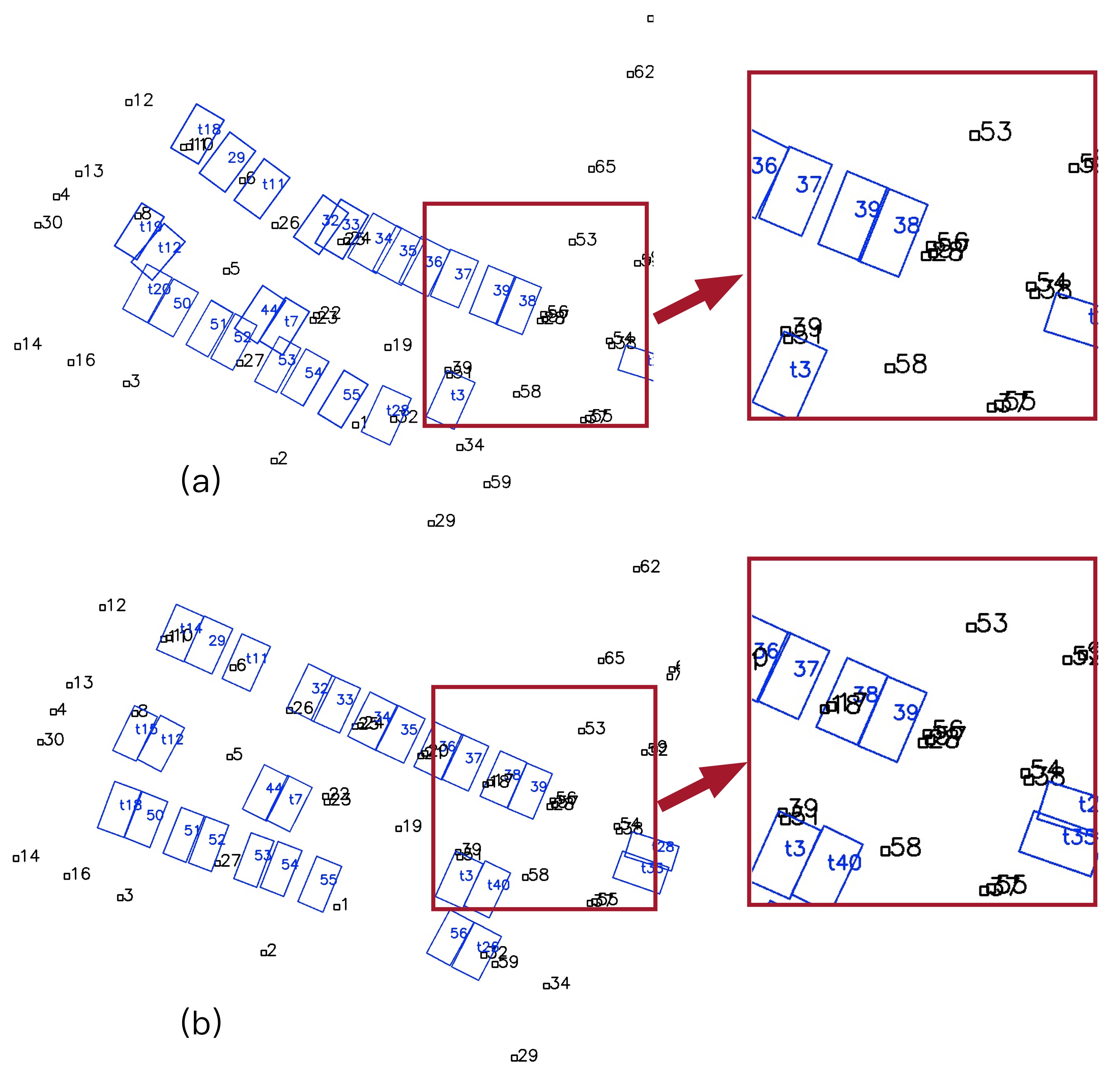}
\centering
\caption{
(a) is the map optimized by traditional graph method without back-end error detection strategy, (b) gives the map result optimized by modified Max-Mixture method.
}\label{map 1}
\end{figure}

In this section, we test our method both online and offline. 
All the parking lot datasets are collected and tested by TiEV autonomous vehicle\footnote{cs1.tongji.edu.cn/tiev}.
We choose a parking lot, who has an area of over 3000 square meters in Jiading Campus, Tongji University, as our test parking lot.
% Since this is an underground parking space, our system experienced harsh condition here.
The dim lighting in the parking lot largely reduces images' quality, while too much light coming from the entry make some of the images overexposed.
Other intruders included pedestrians and vehicles passed by.

Since GPS-based localization methods fail to work correctly in the indoor parking lot, the ground truth is not available.
The standard deviation of the automatically repeated vehicle traces is 0.3m, which is considered as the system accuracy in our situation. 

\subsection{Offline mapping test}
In the offline experiment, several datasets with different starting points are collected. 
Each dataset is processed independently.
The covariance of each observation shifts from 0.1 to 0.25 according to its category and confidence level.
%\COMMENT{When referencing to a figure, should reference exactly, \eg{\Reffig{map 1} (a)}}
\Reffig{map 2} shows the map results of the experimental area.
It maps an extra area of the parking lot without the auxiliary fiducial tags as shown by the red box.
All the parking slots are successfully mapped to the occupied areas of warehouses and staircases represented by vacancies.
% Map blocks with fiducial tags have higher precision as fiducial tags are ideal landmarks which will seldom be wrongly associated.
The undetected parking IDs affect little of the mapping result, as shown in \Reffig{map 2}, by the slots with IDs starting with "t."
We compared the mapping results with and without robust method; the result is shown in \Reffig{map 1}.
In \Reffig{map 1}(a), some slots are wrongly associated with others due to the wrongly detected slot IDs, causing a global ambiguity. \eg{slot No.39 is wrongly recognized as No.38 when firstly identified, and the ID of two slots reversed}. 
In \Reffig{map 1}(b), robust method avoided this from happening by choosing a more reasonable hypothesis.

%

%In the fiducial tag-less area shown in \Reffig{map 2}, 
%\COMMENT{not sure about the meanings}
% wrong association data always occurs.
% However, once a loop is detected, all landmark-positions are corrected.
% Once these wrongly associated slots are corrected, they pile up with one another, which is quite a confusion for the vehicle, and the repeated records of the same landmark is computation-consuming.
%a nearest neighbor search is triggered once loop closure is detected, and duplicate landmarks are merged (\Reffig{map 2}).

\subsection{Online real-time mapping and localization}

\begin{figure}[htbp]
\includegraphics[height = 1.4in]{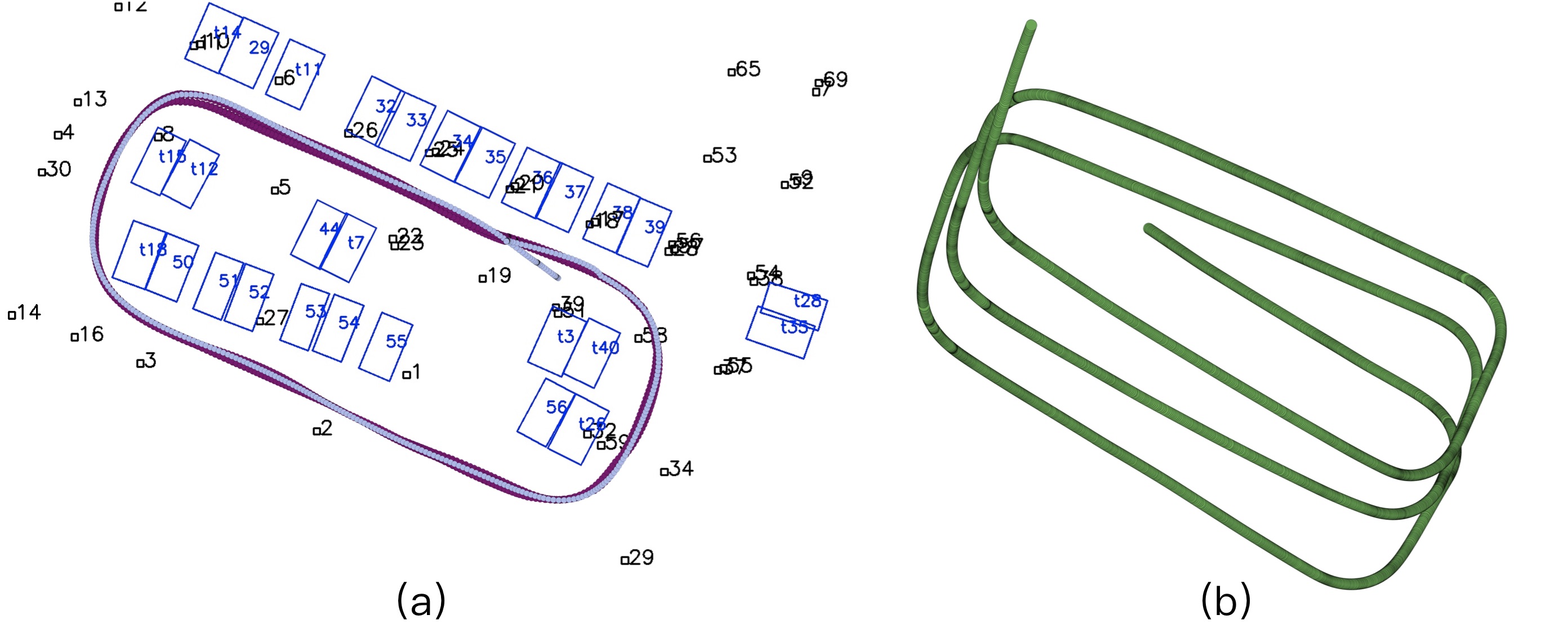}
\caption{
(a) gives a comparison between the human-driving and automatic driving trace, where the nattier blue trace is a human driving trace and the claret trace is the automatic driving trace. (b) is a similar trace recorded using only IMU data. 
}\label{trace compare}
\end{figure}

During the online experiment, the vehicle is first operated by a human driver to initialize the parking map. 
Once the map stabilizes, a car trace is recorded.
Then the vehicle drives automatically at the speed of 3-5 km/h following this pre-recorded trace 
%and fixes its direction constantly 
according to the real-time localization.
The frequency of online part is 10Hz.
The automatic driving procedure is repeated more than ten times; traces are also recorded and compared with the pre-recorded one.
\Reffig{trace compare} shows the traces of both manual (nattier blue) and automatic driving trace (claret).

During the experiment, quite a number of fiducial tags (60 tags) are used to cover the lot-free parts near the entrance and to guarantee a sufficiently stable and credible localization result.

Each tag is printed on an A2-size paper with 48.8 cm side length.
While observing, those tags which are 20 meters or farther than the vehicle are discarded since the accuracy decreases as tags become smaller or even unreadable in the image.
These fiducial tags enable the vehicle to pass through the 3 meters' wide entrance and the long corridor without slots nearby, hence, ensure the robust localization performance.
%What's more, observations with dip-angle are reliable constraints as long as they are within 20 meters.
%\Reffig{}shows the localization result of various tag-vehicle distances.

\enlargethispage{-0.2in}

\section{CONCLUSION}
Due to the various illumination conditions, parking slot is a harsh environment for most SLAM method.
We detect the semantic landmark, parking slots with IDs, in a parking lot, and build the semantic parking incrementally.
In this procedure, semantic data association is vital.
To associate all the semantic information robustly, a robust method for pose graph, Max-Mixture, is utilized and improved.
Experiment in parking lot shows the effectiveness of our system.
However, we use fiducial tags as an aid for loop closure, which is not practical in many circumstances.
In the future work, we aim to replace fiducial tags with other semantic clues including instruction arrows or parking signs on the pillars and improve the adaptability of our system.

\addtolength{\textheight}{-5cm}   % This command serves to balance the column lengths
                                  % on the last page of the document manually. It shortens
                                  % the textheight of the last page by a suitable amount.
                                  % This command does not take effect until the next page
                                  % so it should come on the page before the last. Make
                                  % sure that you do not shorten the textheight too much.

%%%%%%%%%%%%%%%%%%%%%%%%%%%%%%%%%%%%%%%%%%%%%%%%%%%%%%%%%%%%%%%%%%%%%%%%%%%%%%%%

%%%%%%%%%%%%%%%%%%%%%%%%%%%%%%%%%%%%%%%%%%%%%%%%%%%%%%%%%%%%%%%%%%%%%%%%%%%%%%%%

% %%%%%%%%%%%%%%%%%%%%%%%%%%%%%%%%%%%%%%%%%%%%%%%%%%%%%%%%%%%%%%%%%%%%%%%%%%%%%%%%
% \section*{APPENDIX}

% Appendixes should appear before the acknowledgment.

% \section*{ACKNOWLEDGMENT}

% The preferred spelling of the word ÒacknowledgmentÓ in America is without an ÒeÓ after the ÒgÓ. Avoid the stilted expression, ÒOne of us (R. B. G.) thanks . . .Ó  Instead, try ÒR. B. G. thanksÓ. Put sponsor acknowledgments in the unnumbered footnote on the first page.

%%%%%%%%%%%%%%%%%%%%%%%%%%%%%%%%%%%%%%%%%%%%%%%%%%%%%%%%%%%%%%%%%%%%%%%%%%%%%%%%

\bibliographystyle{IEEEtrans}
\bibliography{IEEEabrv,tiev}

\end{document}